# A Video Analysis Method on Wanfang Dataset via Deep Neural Network

Jinlong Kang, Jiaxiang Zheng, Heng Bai, Xiaoting Xue,
Yang Zhou, Jun Guo*
School of Information Science and Technology,Northwest
University, Xi'an, China
Shaanxi International Joint Research Centre for the
Battery-free Internet of things
kjl4ever@outlook.com

Daguang Gan,
Wanfang Data, Beijing 100038, China.
gandg@wanfangdata.com.cn

## ABSTRACT

The topic of object detection has been largely improved recently, especially with the development of convolutional neural network. However, there still exist a lot of challenging cases, such as small object, compact and dense or highly overlapping object. Existing methods can detect multiple objects wonderfully, but because of the slight changes between frames, the detection effect of the model will become unstable, the detection results may result in dropping or increasing the object. In the pedestrian flow detection task, such phenomenon can not accurately calculate the flow. To solve this problem, in this paper, we describe the new function for real-time multi-object detection in sports competition and pedestrians flow detection in public based on deep learning. Our work is to extract a video clip and solve this frame of clips efficiently. More specifically, our algorithm includes two stages: judge method and optimization method. The judge can set a maximum threshold for better results under the model , the threshold value corresponds to the upper limit of the algorithm with better detection results. The optimization method to solve detection jitter problem. Because of the occurrence of frame hopping in the video, and it will result in the generation of video fragments discontinuity. We use optimization algorithm to get the key value, and then the detection result value of index is replaced by key value to stabilize the change of detection result sequence. Based on the proposed algorithm, we adopt wanfang sports competition dataset as the main test dataset and our own test dataset for YOLOv3-Abnormal Number Version(YOLOv3-ANV), which is 5.4% average improvement compared with existing methods. Also, video above the threshold value can be obtained for further analysis. Spontaneously, our work also can used for pedestrians flow detection and pedestrian alarm tasks. Further more, all the code are publicly available for further research: https://github.com/kangjinlong/yolov3_Abnormal-video.

## 1. INTRODUCTION

Multi-object detection is not only one of the basic tasks to be solved in the field of computer vision, but also the basic task of surveillance camera technology. As the objects in video have different attitudes and often appear to be blocked, their movements are irregular. At the same time, considering the diversity of conditions and scenes such as depth of field, resolution, weather and light of video monitoring, the results of object detection algorithm will directly affect the effect of follow-up tracking, classification, motion recognition and behavior description. The basic task of multi-objective detection is still a very challenging task with great potential and space for improvement.

Sports competition is a planned and highly collaborative multi-athlete team behavior. The understanding and recognition of team behavior is one of the important research problems in the field of computer vision research, which has many applications, such as video monitoring, object video summary, human-computer interaction, player assistance training, match decision assistance, video retrieval and browsing, and digital library organization. Faced with a large number of video data, traditional manual processing and analysis methods require a large amount of manpower and a long time, resulting in data can not be processed in a timely manner, thus seriously reducing its practical application value. Therefore, there is a great demand for real-time identification. How to detect athletes in real time is the problem that needs to be solved to complete the above requirements. Whether it is traditional pattern recognition image detection or cnn-based visual detection, it is usually very difficult to detect small object, compact and dense or highly overlapping object. And for a large number of person, some function can not count the object correctly, we call this issue is count problem.

To address these problems, we propose a framework to deal with the count problem in this paper, and the main contribution is that the judge and the optimization are used for different population densities. In the YOLOv3 model, identification algorithm is used to prevent jitter of detection effect for the population. Besides, We propose a function which can count the object number and number the object in sequence.

## 2. Related Work

The idea of object detection is usually to use sliding window detector to extract features through all sliding Windows and then transmit them to classifier, but this method has great computational complexity challenges. Region CNN(**RCNN**)[5] employs the fine-tuned training classification model, implements the idea of Region Proposals to modify the extracted candidate boxes in images so as to make them fit into the input of CNN, and optimizes the candidate boxes with the help of regression. A remarkable effect has been achieved in VOC2007. By virtue of CNN's good feature extraction and classification performance, RCNN carries out feature extraction of candidate regions through Region Proposal method, which reduces the algorithm complexity of traditional sliding window method and greatly improves the detection rate. SPP-NET [6] has made substantial improvement on the basis of RCNN. By replacing the last pooling layer before the full connection layer with the space pyramid pool, the problem of repeated feature extraction calculation of RCNN is effectively solved and the speed bottleneck is broken. Based on the idea of block compatibility features, spp-net connects to the network layer before the full connection layer of fixed input and splits features to solve the problem of fixed input required by CNN. However, SPP-NET still has the problems of multi-stage training and large cost. Fast-RCNN [7] refers to SPP and proposes feature vectors of pooling layer mapping based on region of interest segmentation, so that features of fixed dimensions can be extracted from each region of the image and the overall network training problem existing in SPP-NET can be effectively solved. fast-rcnn realizes real-time end-to-end joint training by establishing multi-task model and using neural network to classify operations. Meanwhile, fast-rcnn implements network terminal synchronization training to improve the accuracy, but there is no significant improvement in the performance of classification steps.

---

*Corresponding author：Jun Guo

Faster-RCNN[8] added Region Proposal Network (RPN) on the basis of fast-rcnn, extracted candidate boxes and merged them into the deep neural Network. Through alternating training, a unified deep neural network framework was established to reduce repeated computation and greatly improve the running speed, almost achieving the optimal effect. Faster RCNN has been put forward for more than two years, but the RPN introduced achieves end-to-end training and generates high-quality regional suggestion boxes, making it still one of the mainstream frameworks in pedestrian detection field. The idea of YOLO[9] is to use a single neural network to directly train the whole input image as the input, so as to more quickly distinguish the background area and object, and conduct real-time monitoring of the object object in a simpler and faster manner. The method divides the input image into S×S size grids, and each grid cell predicts the bounding boxes and the reliability of these bounding boxes. YOLO essentially solves the real-time problem of object detection, and truly realizes the "end-to-end" CNN structure. The idea of YOLOv3[10]'s is, feature maps of a certain size, such as 13*13, are extracted from the input images through the feature extraction network. Then the input images are divided into 13*13 grid cells. If the central coordinates of an object in the ground truth fall in a grid cell, the grid cell will predict the object, because each grid cell predicts a fixed number of bounding boxes. Of these bounding boxes, only those IOU with the maximum limits and the ground truth are used to predict these objects. It can be seen that the predicted output feature map has two dimensions of extracted features.

## 3. YOLOv3-ANV for multi-object detection

Similar to [10], our algorithm changes some aspects, we define it as YOLOv3- Abnormal Number Version(YOLOv3-ANV). Next, I will elaborately introduce the YOLOv3-ANV.

Now there are many popular image detection algorithms and video detection algorithms, all of which have their own features and advantages, and are suitable for use in different situation. Under the YOLOv3 and Faster-RCNN models and population density estimation, we all carry on the counting statistics experiment to the human detection in the general scenes. We find that the YOLOv3 model is more accurate when the number of people is small. The population density is estimated to be an essential test method in a large number of people.

In the case of high density population, population density estimation algorithm is often regard as an effective method to counting the number of people. We use a pixel-based population density estimation algorithm, which uses image and background subtraction to obtain motion prospects, and then linear regression or classification based on foreground pixel area and foreground edge. Among them, the performance of the algorithm is directly affected by the effect of extracting the prospect from the picture. A method of obtaining Adaptive background by Video frame difference method: In video, people are always moving, making the background appear. This method uses the change information of each pixel along the time axis to generate the background image without the target population.

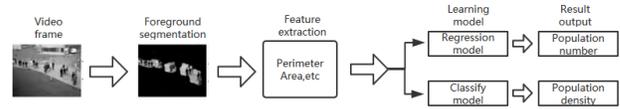

**Figure 1. Process of pixel based method**

In real life, abnormal video clips can be saved as data and used for playback to verify the situation at that time, which is of great significance. Our goal is to extract a video clip that is out of the threshold range of the number of exception in the video clip. In order to achieve this goal, we need to get the exact number of frames in the frame sequence pictures. We tested the model of YOLOv3[10] in the school. Firstly, we run the program to read the collected video data, the program will detecting the frame rate of the video and get the sequence of frames extracted from the video, we save these frames and the result of frame rate in our local. and then use the trained classifier(in the training data set that has been marked ) to recognition the image data (from sequence frames of video), and finally get a series of tagged areas (Bouding-Box regression identified as human areas).Secondly, by counting the number of tagged areas, we can get the number of people in the corresponding image, and according to the "abnormal number threshold" we set to determine whether the threshold is exceeded, so that the video of the abnormal number segment can be intercepted from the original video. Finally, according to the detection jitter prevention method, the recognition results will be optimized. We combine the recognition results of the original image frame as tags into the image, and then use the FFMPEG program (FFmpeg is an open source computer program that can be used to record, convert digital audio, video, and convert it into streams.) to regenerate the video. The generated video corresponds to a series of number of tag frames. At the same time, we will upload the generated information which has the number of person in the image to the remote system for our display system for analysis and display.

Even if YOLOv3[10] has a good effect on the detection of small objects(include person),there will be some leakage detection in the single-frame picture. Because model is limited by the precision of picture itself. In response to this situation, we will set a maximum threshold N for better results under the model(in our experiment, the value of the N is 25, and the accuracy of the model can reach more than 90% in the case of about 20 people). Once the result of using the model to detection a frame is greater than the threshold N, it will be detected by the method of population density estimation[2,3,4], so that the effect is more accurate.

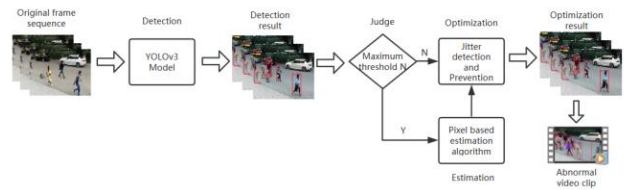

**Figure 2. Process of YOLOv3-ANV method**

Because the model itself has certain generalization ability, the YOLOv3 [10] model trained with training data can detect the population in a variety of scenes, but because of the slight changes between frames, The detection effect of the model will become unstable. We call this unstable change: detection jitter. The detection jitter will reduce the accuracy of collective detection of continuously changing pictures (reference DME calculation index, or custom AP calculation index), and will result in the generation of abnormal video fragments discontinuity, because of the occurrence of frame hopping in the video.

Suppose we find a frame sequence number that is different from that before and after the Detection result, identify it as $x$, and customize the range of an error correction region(length), in our experiment, length takes 1/3 of the video frame rate. It can be expressed by formula (1):

$$length = \frac{1}{3}(N) \tag{1}$$

where $N$ represents the video frame rate. And the detection results of the length range before and after are taken out and counted statistically. It can be expressed by formula(2):

$$count_{x_1} = Count(x-length, x+length, x_1)$$
$$count_{x_2} = Count(x-length, x+length, x_2)$$
$$\ldots \tag{2}$$
$$count_{x_n} = Count(x-length, x+length, x_n)$$

Where $Count()$ represents a function of count, $x-length, x+length$ represent the range of count. $x_1, x_2, \ldots, x_n$ represent the various detection results in the range.

The maximum value $r$ is obtained, and then the detection result value of index $x$ is replaced by $r$ to stabilize the change of detection result sequence! At the same time, the stable abnormal video segment in the range of outside the threshold can also be obtained by using the frame sequence with stable variation of the result. It can be expressed by formula (3):

$$r = \max(count_{x_1}, count_{x2}, \ldots, count_{x_n}) \tag{3}$$

---

**Algorithm** 1: Processing detection result frame by jitter detection

INPUT: The result of detection model

OUTPUT: The optimized detection result

x=0;

DR;// the sequence of detection result

Length=Len(DR);// the length of DR

Last_value;// the Tag of the previous value

$count_{x_n}$ ;// from the Formula 2

length;// 1/3 of video frame rate

**While** x< Length **do**
  **if** x==1 **then**
    Last_value==DR[x];
    x++;
    **continue**;
  **else**
    **if** DR[x]!= last_value **then**
      **for**(i=max(0,x-length); i<x+length; i++) **do**
        **if** (DR[i]==x1) **then**
          $count_{x_1}$ ++;
      **end**
      r=max($count_{x_1}$, $count_{x_2}$, ... , $count_{x_n}$)
      DR[x]=r;
      last_value=r;
    **end**
    x++;
  **end**
**end**

## 4. Experiment
### 4.1 Datasets

In this paper, we adopt wanfang Sports competition data set as the main test dataset for YOLOv3-ANV. It consists of 7 hours 16 minutes sports video. We use the training set Microsoft COCO[11] to train YOLOv3-ANV. We also adopt Experiment with pedestrians on NWU campus, under the surveillance cameras dismount the pedestrians on the road, under the surveillance camera company downstairs and pedestrians on the subway under the surveillance camera data set as our own test dataset for YOLOv3-ANV. In summary, our wanfang datasets are four-fold as follows:

- 2016 American college students's National Rugby League: It includes 3 hours, 28 minutes and 24 seconds of rugby competition.

- 2017 national hockey league : It includes 1 hours, 40 minutes and 41 seconds of hockey competition.

- 2017 ncaa basketball tournament: It includes 1 hours, 54 minutes and 57 seconds of basketball competition.

- The horse racing: It includes 4 minutes and 10 seconds of horse racing.

In addition, our own test datasets are four-fold as follows:

- Experiment with pedestrians on NWU campus：Downstairs, school of information science and technology, northwestern university. It includes 16 minutes and 20 seconds of pedestrians.

- under the surveillance cameras dismount the pedestrians on the road: It includes 1 minutes and 2 seconds of pedestrians.

- under the surveillance camera company downstairs: It includes 31 seconds of pedestrians.

- pedestrians on the subway under the surveillance camera: It includes 31 seconds of pedestrians.

## 4.2 Test wanfang sports competition dataset

### 4.2.1 2016 American college students's National Rugby League

Since the 1960s, American football has surpassed basketball and baseball as America's favorite sport. 1 ~ 2 month of a year, two of the AFC and NFC champions will be in a given city for the Vince Lombardi Trophy (Vince Lombardi Trophy), the total championship is the "Super Bowl" (to use, in the United States football stadium is commonly known as "Bowl"), with more than half of American families TV ratings, more than 150 countries around the world at the same time the game on TV. "Super Bowl Sunday has become the show of the year, basically an unofficial holiday, and it's also the most-watched TV sports show in the country. A number of rugby players have been detected in figure 3.

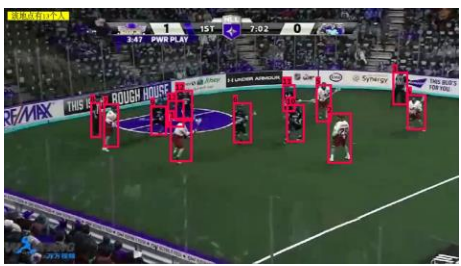

**Figure 3. rugby players of National Rugby League.**

### 4.2.2 2017 National Hockey League

Hockey is an ancient sport, widely developed in the international ball games. It's thousands of years old. Ancient hockey pioneer name who, although has not been able to prove, but can be sure that 3,000 years ago, China, Persia and India and other Asian people love. According to the book "hockey" published by Berlin sports press in 1981, it is known that Chinese soldiers played with sticks and balls 2697 years ago. Images similar to modern hockey games are found on Egyptian pyramids and ancient Greek wall carvings. China in the tang dynasty on the popular "step play" (game time the two teams, and the players hold bottom bend wooden foot to hit the ball, to blow into the bulls win the other goal), its movements and also similar to modern hockey, modern hockey originated in the UK, in the UK in 1861 the first hockey club, in 1875 the British set up the first hockey association, the 1889 men hockey competition held in London. Since then hockey gradually spread to the commonwealth countries. A number of hockey players have been detected in figure 4.

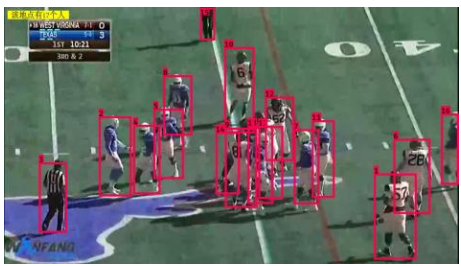

**Figure 4. hockey players of National Hockey League.**

### 4.2.3 2017 NCAA basketball tournament

The National Collegiate Athletic Association is an Association of thousands of colleges and universities in the United States. Its main activities are various sports league held every year, among which the most attention is the basketball league in the first half of the year. Within the NCAA, more than a thousand four-year colleges in the United States and Canada are divided into three grades and dozens of leagues. They play football, basketball, baseball, ice hockey and other sports, including track and field, gymnastics, and wrestling. The NCAA tournament is one of the most important after-school events for American college students. A number of basketball players have been detected in figure 5.

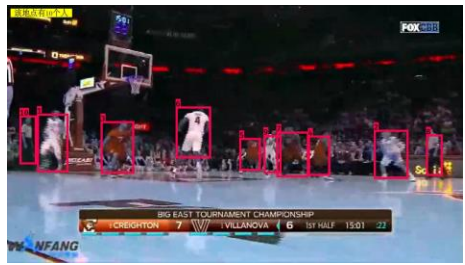

**Figure 5. basketball players of ncaa basketball tournament.**

### 4.2.4 The horse racing

Horse racing is one of the oldest sports. Form has changed a lot since ancient times, but the basic principle is race speed. But horse racing, which attracts spectators like modern horse racing, began in ancient Greece and Rome. In the heyday of the Roman empire there were horse RACES, horse RACES, and what was called Romanesque horse racing, in which riders straddled two horses. Speed horse racing is a kind of competitive activity which is faster than the horse running speed and the ability of the rider to control the horse. Speed racing has a long history. According to research, in the 7th century BC in ancient Greece held in the ancient Olympic Games, there were four horse racing competition, more than 40 years later, the horse racing changed to be controlled by riders. It was originally intended as a means of breeding good horses, only those that did well on the racetrack could be bred. A number of horse racing players have been detected in figure 6.

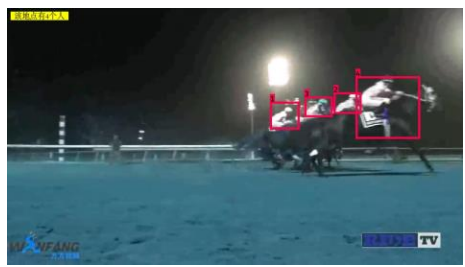

**Figure 6. horse racing players of The horse racing.**

## 4.3 Result of wanfang sports competition dataset

We test the wanfang sports competition dataset in our lab. The GPU is nivdia 1080Ti and the operating system is Ubuntu 16.04 LTS. We use the Faster-RCNN, yolov3 and our model (yolov3-ANV) to test all wanfang sports competition dataset. Detection accuracy has been improved. In order to reflect the detection average precision better, we use some function to compute, and the result can be expressed by formula (4):

$$AP_d = \frac{\sum m1 p1}{\sum m2 p2} \quad (4)$$

where m1 represent the number of detected frames which have the same objects number, m2 represent the number of real frames which have the same objects number, p1,p2 represent the number of detected objects and the number of real objects, and $AP_d$ is the average precision of detection. The result which in Table 1 is the $AP_d$ value.

**Table 1. The evaluation accuracy of different methods for each sports.**

| method | rugby competition | hockey competition | basketball competition | horse racing |
|---|---|---|---|---|
| Faster-RCNN | 66.64% | 60.45% | 64.73% | 64.86% |
| YOLOv3 | 71.57% | 65.16% | 66.75% | 72.28% |
| YOLOv3-ANV | **80.71%** | **75.89%** | **70.46%** | **75.98%** |

## 4.4 Our own test dataset

### 4.4.1 Experiment with pedestrians on NWU campus

To test our algorithm, we took images of our students upstairs in the lab, ready for lunch. A number of students have been detected in figure 7.

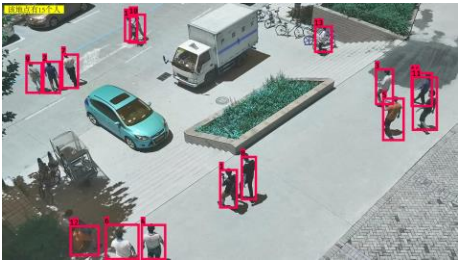

**Figure 7. students of NWU campus.**

#### 4.4.1.1 Under the surveillance cameras dismount the pedestrians on the road

Pedestrians on city road. A number of pedestrians have been detected in figure 8.

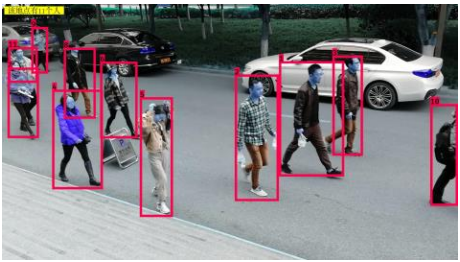

**Figure 8. Pedestrians of surveillance cameras.**

### 4.4.2 Under the surveillance camera company downstairs

Pedestrians on company downstairs. A number of pedestrians have been detected in figure 9.

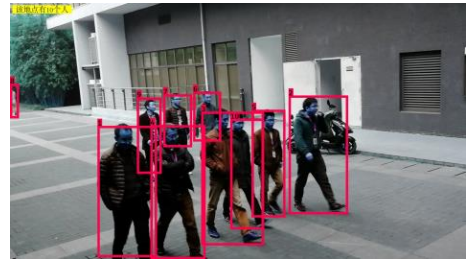

**Figure 9. company downstairs pedestrians of surveillance cameras.**

### 4.4.3 Pedestrians on the subway under the surveillance camera

In public, whether it is a transportation hub, large supermarket, scenic area, the subway, traffic scene change to scenes of public security is a valuable reference index, under the actual constraints of security resources to a large number of configuration, policymakers need to quickly get the scene scene safety rating, a line of information to analysis and early warning system for dangerous scenes for quick necessary guidance.

In the traditional man-flow early-warning mode, the first step is to take the initiative to find out according to the on-site monitoring or the crowd gives feedback from the bottom up, and then the experts judge the situation on the scene, and finally dispatch the police to provide practical guarantee. In this mode, accidents are very likely to occur at the first moment when the crowd is crowded. Attention should be paid to the decision to send personnel to contact and select appropriate security personnel, which has consumed a lot of precious time in the process.

If use the brain to the city idea, will be scattered in different corners of the city monitoring scene data, through the analysis of the terminal of large amounts of data and integration, to easily outbreak crowded scenes and exceed the threshold alarm information management scenarios do real-time analysis, command, and transferring, can grasp the key time node, realize accurate analysis of the traffic supervision task, agile, fast to send, will be treated as possible danger in, to better safeguard the life property safety of the public people. A number of subway flow have been detected in figure 10.

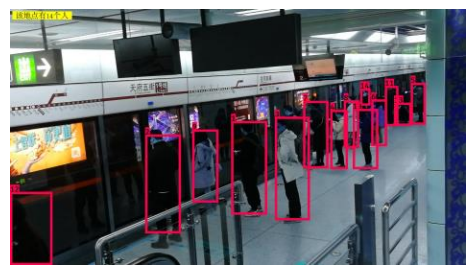

**Figure 10. pedestrians on the subway of surveillance cameras.**

## 4.5 Result of our own test dataset

We also test our own test dataset in our lab.The GPU details is same to 4.2 experiment.We use the Faster-RCNN,yolov3 and our model (yolov3-ANV) to test our own test dataset. Detection accuracy has been improved. The result which in Table 2 is the $AP_d$ value.

**Table 2. The evaluation accuracy of different methods for different scenes.**

| method | students of NWU | road pedestrians | company downstairs | subway |
|---|---|---|---|---|
| Faster-RCNN | 79.34% | 63.65% | 64.73% | 83.28% |
| YOLOv3 | 87.68% | 72.82% | 71.29% | 90.59% |
| YOLOv3-ANV | **90.11%** | **79.49%** | **76.75%** | **91.98%** |

## 5. Conclusion

In this paper, we put forward YOLOv3-ANV, which has Multi-estimation mechanism and detection jitter prevention mechanism, has good application in object detection. Such as multi-player detection in sports competition and public area(campus, city road, company downstairs and subway). Simultaneously, We did some experiments and the experimental results show that YOLOv3-ANV has much better evaluation accuracy for multi-player detection in sports competition and public area than other existing methods. However, we need to further augment data, and the specific images and videos recorded by high resoultion cameras may be much more helpful to our task. Last but not the least, our work also can used for pedestrian flow detection and pedestrian alarm tasks.

## 6. Acknowledgement

This work was supported in part by the National Key Research and Development Program of China under Grant No. 2017YFB1400301. The National Natural Science Foundation of China under grant agreements: Nos. 61973250, 61971349, 61973249, 61702415. The ShaanXi Science and Technology Innovation Team Support Project under grant agreement 2018TD-026. Talent Support Project of Science Association in Shaanxi Province: 20180108. Scientific research plan for servicing local area of Shaanxi province education department: 19JC038, 19JC041.